\documentclass[runningheads]{llncs}
\usepackage{graphicx}
\usepackage{amsmath,amssymb} 
\usepackage{color}
\usepackage{algorithm}
\usepackage{algorithmic}
\usepackage{subfig}
\usepackage{authblk}

\begin{document}
\title{Incomplete Multi-view Clustering via Graph Regularized Matrix Factorization}

\titlerunning{Incomplete Multi-view Clustering}

\author{Jie Wen\inst{1} \thanks{ $\star$ indicates equal contributions;  $\dagger$ indicates the corresponding author.}  \and
Zheng Zhang\inst{1,2}$^{\star}$  \and
Yong Xu\inst{1}$^{\dagger}$ \and Zuofeng Zhong \inst{1}}
\authorrunning{Jie Wen, Zheng Zhang, and Yong Xu et al.}

\institute{Bio-Computing Research Center, Harbin Institute of Technology, Shenzhen, Shenzhen 518055, Guangdong, China\\
\and
The University of Queensland, Australia\\
\email{wenjie@hrbeu.edu.cn; darrenzz219@gmail.com; yongxu@ymail.com; zfzhong2010@gmail.com}}

%
%

\maketitle              
\begin{abstract}
Clustering with incomplete views is a challenge in multi-view clustering. In this paper, we provide a novel and simple method to address this issue. Specially, the proposed method simultaneously exploits the local information of each view and the complementary information among views to learn the common latent representation for all samples, which can greatly improve the compactness and discriminability of the obtained representation. Compared with the conventional graph embedding methods, the proposed method does not introduce any extra regularization term and corresponding penalty parameter to preserve the local structure of data, and thus does not increase the burden of extra parameter selection. By imposing the orthogonal constraint on the basis matrix of each view, the proposed method is able to handle the out-of-sample. Moreover, the proposed method can be viewed as a unified framework for multi-view learning since it can handle both incomplete and complete multi-view clustering and classification tasks. Extensive experiments conducted on several multi-view datasets prove that the proposed method can significantly improve the clustering performance.

\end{abstract}
\section{Introduction}
Multi-view clustering has been achieved great development and has been successfully applied in many applications, such as image retrieval \cite{li2009exploiting}, webpage classification \cite{blum1998combining,Zhang2018Highly}, and speech recognition \cite{ngiam2011multimodal}. Recently, many methods have been proposed, such as multi-view \emph{k}-means clustering \cite{cai2013multi}, multi-view spectral clustering via bipartite graph \cite{li2015large}, and co-regularized multi-view spectral clustering \cite{kumar2011co}, etc. Compared with the single-view clustering, multi-view clustering can exploit the complementary information among multiple views, and thus has the potential to achieve a better performance \cite{zhao2017multi}.

For the conventional multi-view clustering, they commonly require that the available samples should have all of the views. However, it always happens that some views are missing for parts of samples in real world applications \cite{trivedi2010multiview}. For example, the data obtained by the blood test and images scanned by the magnetic resonance can be regarded as two necessary views for diagnosing the disease. However, it is often the case that we only have the results of one view for some individuals since they would like to take only one of the two tests. In this case, the conventional methods fail. In this paper, we refer to the clustering task with incomplete views as incomplete multi-view clustering (IMC).

For IMC, a few methods have been proposed, which can be commonly categorized into two groups. The one group is based on completing the incomplete views. For example, Trivedi et al. proposed a kernel CCA based method, which tries to recover the kernel matrix of the incomplete view and then learns two projections for the two views, respectively \cite{trivedi2010multiview}. However, it requires at least one complete view for reference. In other words, it is not applicable to the case that all views are incomplete. To address this issue, Gao et al. proposed a two-step approach, which first fills in the missing views with the corresponding average of all samples, and then learns the common representation for the two views based on the spectral graph theory \cite{gao2016incomplete}. The shortcoming of this approach is that it introduces some useless even noisy information to the data. For data with small incomplete percentages, this approach may be effective. However, for the data with large incomplete percentages, this approach is harmful to find the common representation since these useless information may dominate the representation learning \cite{shao2015multiple}. The other group focuses on directly learning the common latent subspace or representation for all views, in which the most representative works are the partial multi-view clustering (PVC) \cite{zhi2014partial}, multi-incomplete-view clustering (MIC) \cite{shao2015multiple}, and incomplete multi-modality grouping (IMG) \cite{zhao2016incomplete}. Based on the non-negative matrix factorization (NMF), PVC directly learns a common latent representation for two views by simply regularizing different views of the same sample to have the same representation \cite{zhi2014partial}. MIC jointly learns the latent representation of each view and the consensus representation by utilizing the weighted NMF algorithm, in which the missing views are constrained with the small weight even 0 during learning \cite{shao2015multiple}. IMG can be viewed as an extension of PVC, which further embeds an adaptively learned graph on the latent representation \cite{zhao2016incomplete}.

Although some methods have been proposed to address the IMC problem, several problems still exist which limit their performances. For example, these methods all ignore the geometric structure of data. This indicates that the intrinsic geometric structure of data may be destroyed in the representation space, which may lead to a bad performance. The second shortcoming of these methods, especially MIC and IMG, is that there are many penalty parameters (more than three) to be set. These tunable parameters directly influence the clustering performance and limit its real applications because it is still an open problem to adaptively select the optimal parameter for different datasets. The third shortcoming is that these methods all cannot handle the out-of-sample problem. In this paper, we propose a novel and simple IMC method, named incomplete multi-view clustering via graph regularized matrix factorization (IMC\_GRMF), to solve the above problems and improve the performance. Similar to PVC, the matrix factorization technique is exploited to learn the common latent representation, in which the representation corresponding to those samples with all views are regularized to be consistent. In addition, a nearest neighbor graph is neatly imposed on the reconstruction errors of the matrix factorization to exploit the local geometric structure of data, which enables the method to learn a more compact and discriminative representation for clustering. Compared with the other methods, our approach does not introduce any extra regularization term and corresponding penalty parameter to preserve the locality structure of data. Extensive experimental results prove the effectiveness of the proposed method for incomplete multi-view clustering.

\section{Notations and related work}
\subsection{Notations}
Let ${X^{( k )}} = {[ {X_c^{\left( k \right)T};{{\bar X}^{\left( k \right)T}}}]^T} \in {R^{( {{n_c} + {n_k}}) \times {m_k}}}$ be the $k$th view of data, where each sample in the corresponding view is represented by a row vector with ${m_k}$ features, ${n_c}$ is the number of paired samples (\emph{i.e.}, there are no missing views for these samples). $x_i^{(k)}$ denotes the features of the $k$th view of the $i$th sample. We refer to the $k$th view as $Vi\left( k \right)$. ${\bar X^{\left( k \right)}} \in {R^{{n_k} \times {m_k}}}$ represents that ${n_k}$ samples only contain the features of $Vi\left( k \right)$ while the features of the other views are missing. The total samples of the data is $n = {n_c} + \sum\limits_{k = 1}^v {{n_k}} $. For a matrix $A \in {R^{m \times n}}$, its ${l_F}$ norm and ${l_1}$ norm are defined as ${\left\| A \right\|_F} = \sqrt {\sum\limits_{j = 1}^n {\sum\limits_{i = 1}^m {a_{i,j}^2} } } $ and ${\left\| A \right\|_1} = \sum\limits_{j = 1}^n {\sum\limits_{i = 1}^m {\left| {{a_{i,j}}} \right|} } $, respectively, where ${a_{i,j}}$ denotes the $i$th row and $j$th column element of matrix $A$ \cite{zhang2013binary,qin2017binary}. $Tr\left(  \cdot  \right)$ is the trace operation. We use ${A^T}$ to denote the transposition of matrix $A$ \cite{qin2017zero}. $I$ is the identity matrix. $A \ge 0$ means that all elements of matrix $A$ are not less than zero.

\subsection{Partial multi-view clustering (PVC)}
For data with two incomplete views, PVC seeks to learn a common latent subspace for both two views, where different views of the same sample should have the same representation [14]. The learning model of PVC is formulated as follows:
\begin{equation}\label{1}
\begin{array}{r}
\mathop {\min }\limits_{{P_c},{{\bar P}^{(1)}},{{\bar P}^{(2)}},{U^{(1)}},{U^{(2)}}} \left\| {\left[ \begin{array}{l}
X_c^{(1)}\\
{{\bar X}^{(1)}}
\end{array} \right] - \left[ \begin{array}{l}
{P_c}\\
{{\bar P}^{(1)}}
\end{array} \right]{U^{(1)}}} \right\|_F^2 + \lambda {\left\| {\left[ \begin{array}{l}
{P_c}\\
{{\bar P}^{(1)}}
\end{array} \right]} \right\|_1}\\
 + \left\| {\left[ \begin{array}{l}
X_c^{(2)}\\
{{\bar X}^{(2)}}
\end{array} \right] - \left[ \begin{array}{l}
{P_c}\\
{{\bar P}^{(2)}}
\end{array} \right]{U^{(2)}}} \right\|_F^2 + \lambda {\left\| {\left[ \begin{array}{l}
{P_c}\\
{{\bar P}^{(2)}}
\end{array} \right]} \right\|_1}\\
s.t.{\kern 1pt} {\kern 1pt} {U^{(1)}} \ge 0,{U^{(2)}} \ge 0,{P_c} \ge 0,{{\bar P}^{(1)}} \ge 0,{{\bar P}^{(2)}} \ge 0,
\end{array}
\end{equation}
where $\lambda $ is the penalty parameter. ${U^{(1)}} \in {R^{K \times {m_1}}}$ and ${U^{(2)}} \in {R^{K \times {m_2}}}$ are the latent space basis matrices for the two views, ${P_c} \in {R^{{n_c} \times K}}$, ${\bar P^{(1)}} \in {R^{{n_1} \times K}}$, and ${\bar P^{(2)}} \in {R^{{n_2} \times K}}$ are the latent representations of the original data, $K$ is the feature dimension in the latent space.

For PVC, the new representation corresponding to all samples can be expressed as $P = \left[ \begin{array}{l}{P_c}\\{{\bar P}^{(1)}}\\{{\bar P}^{(2)}}\end{array} \right] \in {R^{n \times K}}$. Then the conventional \emph{k}-means can be performed on it to obtain the final clustering result.

\section{The proposed method}
For multi-view data, learning a common latent representation for all views is one of the most favorite approaches in the field of multi-view clustering. However, how to learn a compact and discriminative common representation for the incomplete multi-view data is a challenge task. In this section, a novel multi-view clustering framework shown in Fig.1 is provided to address this issue, in which the local information of each view and the complementary information across different views are jointly integrated.

\begin{figure*}[htbp!]
\centering
\centering\resizebox{.96\textwidth}{.25\textheight}{
\includegraphics[]{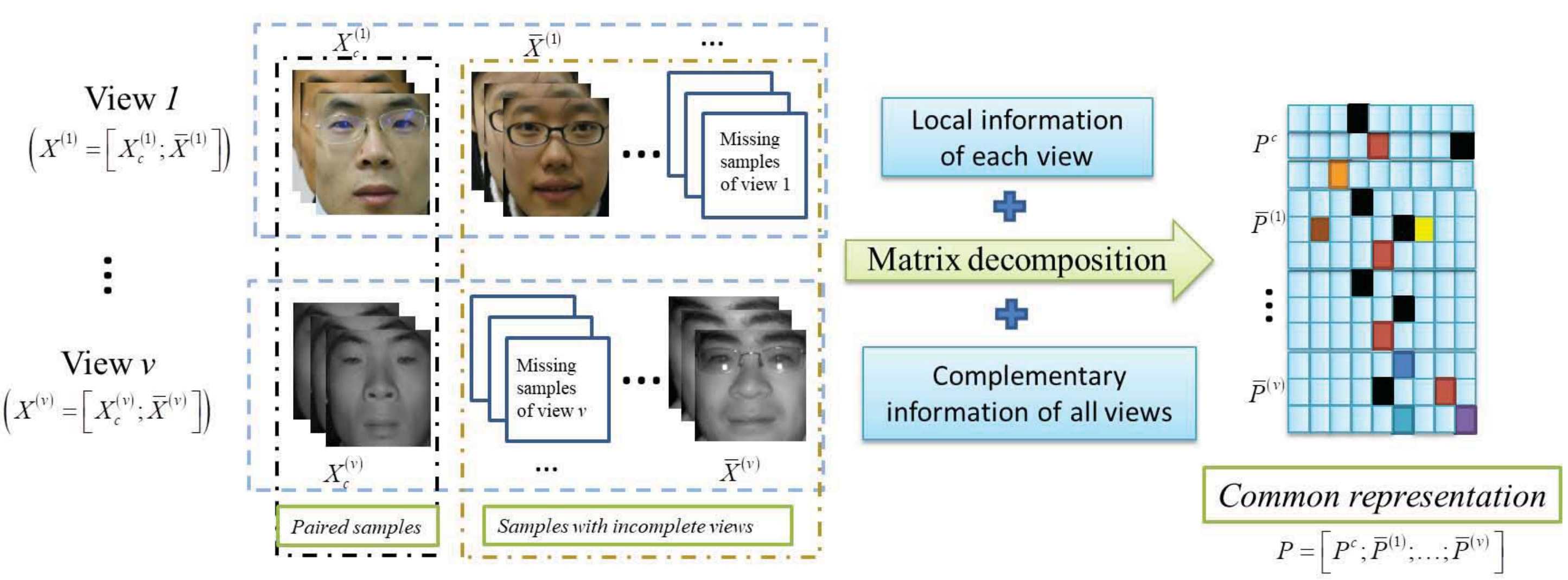}}
\caption{The description of IMC\_GRMF. In this work, we suppose that there are only $n_c$ samples (paired samples) have features of all views.}
\label{fig:2}
\end{figure*}

\subsection{Learning model of the proposed method}
In past years, exploiting the locality geometric structure of data has been proved an effective approach for representation learning, which not only can improve the discriminability and compactness of the learned representation, but also avoids overfitting \cite{zhang2017marginal,wen2018lowrank,fei2016low,zhang2018semi,zhang2017discriminative,qian2016double,rai2016partial}. For example, in \cite{qian2016double,rai2016partial}, a nearest neighbor graph is introduced to constrain the new representation or basis for incomplete multi-view clustering. Although the purpose is realized, the complexity is also increased because they commonly introduce at least one tunable penalty parameter to the model. Since some basic models already have two or more tuned parameters, introducing any extra tuned parameter to the model will greatly increase the burden in parameter selection. So the conventional graph embedding approaches are not a good choice to guide the representation learning. In this section, we propose a novel and simple approach to solve this challenge, in which the local information of each view are embedded into the learning model based on the following Lemma \cite{Wen2018Low}.

\emph{\textbf{Lemma 1}}: For three samples $\left\{ {{x_1},{x_2},{x_3}} \right\} \in {R^m}$, suppose ${x_1}$ and ${x_2}$ are the nearest neighbor to each other, ${x_3}$ is not the nearest neighbor to samples ${x_1}$ and ${x_2}$. If there is a complete dictionary $U \in {R^{k \times m}}$ that satisfies ${x_i} = {p_i}U$ ($i = \left\{ {1,2,3} \right\}$), where ${p_i} \in {R^k}$ can be viewed as the reconstruction coefficient. Then we have the following conclusion: the reconstructed sample ${p_2}U$ (${p_1}U$) is also the nearest neighbor to the original sample ${x_1}$ (${x_2}$) and is still not the nearest neighbor to sample ${x_3}$.

The proof to Lemma 1 is very simple and thus we omit it here. From Lemma 1, we know that the reconstruction operation does not destroy the local geometric structure of the original data. Inspired by this motivation, we design the following objective function to exploit the local information of data for common representation learning:
\begin{equation}\label{2}
\begin{array}{r}
\mathop {\min }\limits_{{P^{\left( k \right)}},{U^{\left( k \right)}}} \sum\limits_{k = 1}^v {\sum\limits_{j = 1}^{{n_c} + {n_k}} {\sum\limits_{i = 1}^{{n_c} + {n_k}} {\left\| {x_i^{(k)} - p_j^{(k)}{U^{(k)}}} \right\|_2^2w_{i,j}^{(k)}} } } {\rm{ + }}{\lambda _2}\sum\limits_{k = 1}^v {{{\left\| {{P^{(k)}}} \right\|}_1}} \\
s.t.{U^{(k)}}{{{U^{(k)T}}}} = I,
\end{array}
\end{equation}
where ${\lambda _2}$ is a penalty parameter. $p_j^{\left( k \right)}$ is the new representation of the $j$th sample in the $k$th view. $w_{i,j}^{(k)}$ is a binary value which is simply pre-defined as follows:
\begin{equation}\label{3}
w_{i,j}^{(k)} = \left\{ {\begin{array}{*{20}{c}}
{1,}&{if{\kern 1pt} {\kern 1pt} x_i^{\left( k \right)} \in \Phi \left( {x_j^{\left( k \right)}} \right){\kern 1pt} {\kern 1pt} {\kern 1pt} or{\kern 1pt} {\kern 1pt} x_j^{\left( k \right)} \in \Phi \left( {x_i^{\left( k \right)}} \right)}\\
{0,}&{otherwise},
\end{array}} \right.
\end{equation}
where $\Phi \left( {x_j^{\left( k \right)}} \right)$ denotes the sample set of nearest neighbors to sample $x_j^{\left( k \right)}$.

By introducing the binary weight to regularize the data reconstruction, the locality structure of the original data in each view can be well preserved. Meanwhile, from (2), we can find that the proposed method does not introduce any extra regularization term and corresponding tuned parameter to preserve such locality property, which greatly reduces the complexity of penalty parameter selection in comparison with the other graph regularized IMC methods, such as DCNMF \cite{qian2016double} and GPMVC \cite{rai2016partial} which all commonly introduce at least an extra tuned penalty parameter to preserve such locality property.
For the paired samples across different views, their new representation should be consensus. To this end, we further add a regularization term based on the paired information of different views as follows:
\begin{equation}\label{4}
\begin{array}{r}
\mathop {\min }\limits_{{P^{\left( k \right)}},{P^c},{U^{\left( k \right)}}} \sum\limits_{k = 1}^v {\sum\limits_{j = 1}^{{n_c} + {n_k}} {\sum\limits_{i = 1}^{{n_c} + {n_k}} {\left\| {x_i^{(k)} - p_j^{(k)}{U^{(k)}}} \right\|_2^2w_{i,j}^{(k)}} } } \\
{\rm{ + }}{\lambda _1}\sum\limits_{k = 1}^v {\left\| {{G^{(k)}}{P^{(k)}} - {P^c}} \right\|_F^2}  + {\lambda _2}\sum\limits_{k = 1}^v {{{\left\| {{P^{(k)}}} \right\|}_1}} \\
s.t.{U^{(k)}}{{{U^{(k)T}}}} = I,
\end{array}
\end{equation}
where ${\lambda _1}$ is a penalty parameter. ${P^c} \in {R^{c \times K}}$ is the common latent representation for the paired samples of different views. ${G^{\left( k \right)}} \in {R^{{n_c} \times \left( {{n_c} + {n_k}} \right)}}$ can be viewed as an index matrix used to remove the unpaired representation ${\bar P^{\left( k \right)}}$ from ${P^{\left( k \right)}} = \left[ \begin{array}{l}P_c^{\left( k \right)}\\{{\bar P}^{\left( k \right)}}\end{array} \right]$. Since the first $n_c$ samples of each view are regarded as the paired samples, matrix ${G^{\left( k \right)}}$ can be simply defined as follows:
\begin{equation}\label{5}
G_{i,j}^{\left( k \right)} = \left\{ {\begin{array}{*{20}{c}}
{1,}&{if{\kern 1pt} {\kern 1pt} i = j}\\
{0,}&{otherwise}.
\end{array}} \right.
\end{equation}

For model (4), $P = [P^{cT},{\bar P}^{( 1)T}, \ldots ,{\bar P}^{( v)T}]^T$ can be viewed as the new representations for all samples. After obtaining the new representations, we use \emph{k}-means algorithm to partition those samples into their respective groups. Several good properties of the proposed model (4) are summarized as follows.

\textbf{Remark 1}: The proposed method is not only a clustering algorithm, but also an unsupervised classification algorithm because it can handle the out-of-sample. In essence, for any sample $x_i^{(k)}$ in the $k$th view, its new representation is obtained by the matrix factorization $x_i^{(k)}=p_i^{(k)}U^{(k)}$, which is equivalent to $x_i^{(k)}{U^{\left( k \right)T}}{\rm{ = }}p_i^{(k)}$ since ${U^{(k)}}{{{U^{(k)T}}}} = I$. Therefore, when the basis matrix ${U^{\left( k \right)}}$ is obtained, we can first achieve the discriminative representation for any new coming sample ${y^{\left( k \right)}}$ by projecting it onto the basis matrix as $p_y^{(k)}{\rm{ = }}{{{y}}^{\left( k \right)}}{U^{\left( k \right)T}}$, and then use the conventional unsupervised classification methods like \emph{k}-nearest neighbor classify to predict its label.

\textbf{Remark 2}: The proposed model (4) is a unified multi-view learning framework, which can be applied to the incomplete and complete cases by defining different index matrixes ${G^{(k)}}$.

\textbf{Remark 3}: The proposed method simultaneously exploits the local information of each view and the complementary information across different views, which is beneficial to learn a more compact and discriminative representation for clustering, and thus has the potential to perform better. Moreover, embedding the local information into the model can avoid the overfitting in handing the new sample.

\textbf{Remark 4:} Most importantly, we do not introduce any extra regularization term to preserve the local geometric structure of data. In other words, compared with the conventional graph embedding methods, the proposed method does not increase the burden of parameter tuning.

\textbf{Remark 5:} The proposed method has the potential to recover the missing views. Specifically, for a sample with only the $k$th view ${x^{\left( k \right)}}$, when its new representation ${p_{{x^{\left( k \right)}}}}$ is obtained via the proposed method, we can recover its $f$th missing view via ${x^{\left( f \right)}} = {p_{{x^{\left( k \right)}}}}{U^{\left( f \right)}}$.

\subsection{Solution to IMC\_GRMF}
For the first term of (4), we can rewrite it into the following equivalent formula
\begin{equation}\label{6}
\begin{split}
&\sum\limits_{k = 1}^v {\sum\limits_{j = 1}^{{n_c} + {n_k}} {\sum\limits_{i = 1}^{{n_c} + {n_k}} {\left\| {x_i^{(k)} - p_j^{(k)}{U^{(k)}}} \right\|_2^2w_{i,j}^{(k)}} } } \\
 = &\sum\limits_{k = 1}^v {\left( \begin{array}{l}
Tr\left( {{X^{\left( k \right)T}}{D^{\left( k \right)}}{X^{\left( k \right)}}} \right) + Tr\left( {{U^{\left( k \right)T}}{P^{\left( k \right)T}}{D^{\left( k \right)}}{P^{\left( k \right)}}{U^{\left( k \right)}}} \right)\\
 - 2Tr\left( {{X^{\left( k \right)T}}{W^{\left( k \right)}}{P^{\left( k \right)}}{U^{\left( k \right)}}} \right)
\end{array} \right)},
\end{split}
\end{equation}
where ${D^{\left( k \right)}}$ is a diagonal matrix with each diagonal element $D_{i,i}^{\left( k \right)} = \sum\limits_{j = 1}^{{n_c} + {n_k}} {W_{i,j}^{\left( k \right)}} $. Considering that the first term of (6) is constant and condition ${U^{(k)}}{{{U^{(k)T}}}} = I$, we can simplify (4) as follows according to (6):
\begin{equation}\label{7}
\begin{split}
L\left( {{P^{\left( k \right)}},{P^c},{U^{\left( k \right)}}} \right) = {\lambda _1}\sum\limits_{k = 1}^v {\left\| {{G^{(k)}}{P^{(k)}} - {P^c}} \right\|_F^2}  + {\lambda _2}\sum\limits_{k = 1}^v {{{\left\| {{P^{(k)}}} \right\|}_1}} \\
 + \sum\limits_{k = 1}^v {\left( {Tr\left( {{P^{\left( k \right)T}}{D^{\left( k \right)}}{P^{\left( k \right)}}} \right) - 2Tr\left( {{X^{\left( k \right)T}}{W^{\left( k \right)}}{P^{\left( k \right)}}{U^{\left( k \right)}}} \right)} \right)}.
\end{split}
\end{equation}

Then all variables can be calculated alternatively as follows.

\textbf{Step 1}: Calculate ${U^{\left( k \right)}}$. The basis matrix ${U^{\left( k \right)}}$ for each view can be calculated by optimizing the following problem:
\begin{equation}\label{8}
\begin{split}
\mathop {\min }\limits_{{U^{\left( k \right)}}{U^{\left( k \right)T}} = I}  - 2Tr\left( {{X^{\left( k \right)T}}{W^{\left( k \right)}}{P^{\left( k \right)}}{U^{\left( k \right)}}} \right).
\end{split}
\end{equation}

Then we can obtain the optimum value of ${U^{\left( k \right)}}$ as \cite{zou2006sparse,wen2018robust}:
\begin{equation}\label{9}
{U^{(k)}} = {J^{(k)}}{B^{(k)T}},
\end{equation}
where ${J^{(k)}}$ and ${B^{(k)}}$ are the right and left singular matrices of $({X^{(k)T}}{W^{(k)}}{P^{(k)}})$, \emph{i.e.}, ${X^{(k)T}}{W^{(k)}}{P^{(k)}} = {B^{(k)}}{\Sigma ^{(k)}}{J^{(k)T}}$.

\textbf{Step 2}: Calculate ${P^{\left( k \right)}}$. Fixing the other variables, variable ${P^{\left( k \right)}}$ can be calculated by minimizing the following problem:
\begin{equation}\label{10}
\begin{split}
\mathop {\min }\limits_{{P^{\left( k \right)}}} {\lambda _1}\left\| {{G^{(k)}}{P^{(k)}} - {P^c}} \right\|_F^2 + {\lambda _2}{\left\| {{P^{(k)}}} \right\|_1}\\
 + Tr\left( {{P^{\left( k \right)T}}{D^{\left( k \right)}}{P^{\left( k \right)}}} \right) - 2Tr\left( {{X^{\left( k \right)T}}{W^{\left( k \right)}}{P^{\left( k \right)}}{U^{\left( k \right)}}} \right).
\end{split}
\end{equation}

Define ${A^{\left( k \right)}} = {U^{\left( k \right)}}{X^{\left( k \right)T}}{W^{\left( k \right)}} + {\lambda _1}{P^{cT}}{G^{\left( k \right)}}$, ${M^{\left( k \right)}} = {D^{\left( k \right)}} + {\lambda _1}{G^{\left( k \right)T}}{G^{\left( k \right)}}$. Obviously, ${M^{\left( k \right)}}$ is still a diagonal matrix with all diagonal elements $M_{i,i}^{\left( k \right)} > 0$. Thus, (10) can be rewritten into the following equivalent problem:
\begin{equation}\label{11}
\begin{split}
 \mathop {\min }\limits_{{P^{\left( k \right)}}} \left\| {{{\left( {{M^{\left( k \right)}}} \right)}^{\frac{1}{2}}}{P^{\left( k \right)}} - {{\left( {{A^{\left( k \right)}}{{\left( {{M^{\left( k \right)}}} \right)}^{{\rm{ - }}\frac{1}{2}}}} \right)}^T}} \right\|_F^2 + {\lambda _2}{\left\| {{P^{\left( k \right)}}} \right\|_1}.
\end{split}
\end{equation}

Define ${C^{\left( k \right)}} = ({A^{\left( k \right)}}{\left( {{M^{\left( k \right)}}} \right)^{{\rm{ - }}\frac{1}{2}}})^T$, problem (11) can be rewritten as follows
\begin{equation}\label{13}
\mathop {\min }\limits_{{P^{\left( k \right)}}} \sum\limits_{i = 1}^{{n_c} + {n_k}} {{{M_{i,i}^{\left( k \right)}}}\left\| {P_{i,:}^{\left( k \right)} - {{C_{i,:}^{\left( k \right)}} \mathord{\left/
 {\vphantom {{C_{i,:}^{\left( k \right)}} {\sqrt {M_{i,i}^{\left( k \right)}} }}} \right.
 \kern-\nulldelimiterspace} {\sqrt {M_{i,i}^{\left( k \right)}} }}} \right\|_2^2 + {\lambda _2}{{\left\| {P_{i,:}^{\left( k \right)}} \right\|}_1}}.
\end{equation}
where $P_{i,:}^{\left( k \right)}$ and $C_{i,:}^{\left( k \right)}$ denote the $i$th row vector of matrices ${P^{\left( k \right)}}$ and ${C^{\left( k \right)}}$, respectively. For problem (12), its solution can be computed independently to each row by the conventional shrinkage operation as follows \cite{wen2018robust}:
\begin{equation}\label{14}
P_{i,:}^{\left( k \right)} = {\Theta _{{{{\lambda _2}} \mathord{\left/
 {\vphantom {{{\lambda _2}} {\sqrt[4]{{M_{i,i}^{\left( k \right)}}}}}} \right.
 \kern-\nulldelimiterspace} {2{{M_{i,i}^{\left( k \right)}}}}}}}\left( {{{C_{i,:}^{\left( k \right)}} \mathord{\left/
 {\vphantom {{C_{i,:}^{\left( k \right)}} {\sqrt {M_{i,i}^{\left( k \right)}} }}} \right.
 \kern-\nulldelimiterspace} {\sqrt {M_{i,i}^{\left( k \right)}} }}} \right),
\end{equation}
where $\Theta$ denotes the shrinkage operator.

\textbf{Step 3}: Calculate ${P^c}$. Fixing the other variables, the common latent representation ${P^c}$ can be calculated by solving the following minimization problem:
\begin{equation}\label{14}
\mathop {\min }\limits_{{P^c}} \sum\limits_{k = 1}^v {\left\| {{G^{(k)}}{P^{(k)}} - {P^c}} \right\|_F^2}.
\end{equation}

Problem (14) has the following closed form solution:
\begin{equation}\label{15}
{P^c} = {\sum\limits_{k = 1}^v {{G^{\left( k \right)}}{P^{\left( k \right)}}}}/v.
\end{equation}

Algorithm 1 summarizes the computing procedures of IMC\_GRMF.
\begin{algorithm}[htb]
 \caption{: IMC\_GRMF (solving problem (4))}
  \begin{algorithmic}
  \STATE {\textbf{Input:}  Multi-view ${X^{\left( k \right)}}$, index matrix ${G^{\left( k \right)}}$, $k \in \left[ {1,v} \right]$, parameters  ${\lambda _1}$, ${\lambda _2}$.
   \\\textbf{Initialization:} Initialize ${P^{\left( k \right)}}$ and ${U^{\left( k \right)}}$ with random values, construct the nearest neighbor graph ${W^{\left( k \right)}}$, ${P^c} = {{\sum\limits_{k = 1}^v {{G^{\left( k \right)}}{P^{\left( k \right)}}} } \mathord{\left/
 {\vphantom {{\sum\limits_{k = 1}^v {{G^{\left( k \right)}}{P^{\left( k \right)}}} } v}} \right.
 \kern-\nulldelimiterspace} v}$.}
  \WHILE{not converged}{\STATE
  \textbf{for} $k$ from 1 to $v$ \\
  {\STATE \begin{enumerate}
    \item[] Update ${U^{\left( k \right)}}$ using (9).
    \item[] Update ${P^{\left( k \right)}}$ using (13).
  \end{enumerate}}
  \textbf{end}}\\
Update ${P^c}$ using (15).\\
\ENDWHILE
\STATE {\textbf{Output:} ${P^c},{P^{\left( k \right)}},{U^{\left( k \right)}}$}
\end{algorithmic}
\end{algorithm}

\subsection{Computational complexity and convergence property}
For Algorithm 1, it is obvious that the biggest computational cost is the singular value decomposition (SVD) in Step 1. Note that the computational complexities of matrix multiplication and addition are ignored since their computational costs are far less than SVD. Thus, we only take into account the computational complexity of Step 1. Generally, the computational complexity of SVD is $O({mn^2})$ for a $m \times n$ matrix \cite{liu2013robust}. Therefore, the computational complexity of Step 1 is about $O\left( {v{mK^2}} \right)$. $v$ is the number of views, $K$ is the reduced dimension or the number of clusters. Therefore, the computation complexity of the proposed method listed in Algorithm 1 is about $O\left( {\tau vm{K^2}} \right)$, where $\tau $ is the iteration number.

From the above presentations, it is obvious to see that the proposed optimization problem (7) is convex with respect to variables ${P^{\left( k \right)}},{P^c},{U^{\left( k \right)}}$, respectively. Then we have the following \textbf{Theorem 1}.

\emph{\textbf{Theorem 1}}: The objective function value of problem (4) is monotonically decreasing during the iteration.

\textbf{\emph{Proof}}. Suppose $\Upsilon \left( {P_t^{\left( k \right)},P_t^c,U_t^{\left( k \right)}} \right)$ denotes the objective function value at the $t$th iteration. Since all sub-problems with respect to variables ${P^{\left( k \right)}},{P^c},{U^{\left( k \right)}}$ are convex and have the closed form solution, the following inequations are satisfied:
\begin{equation}\label{16}
\begin{array}{l}
\Upsilon \left( {P_t^{\left( k \right)},P_t^c,U_t^{\left( k \right)}} \right) \ge \Upsilon \left( {P_t^{\left( k \right)},P_t^c,U_{t + 1}^{\left( k \right)}} \right)\\
 \ge \Upsilon \left( {P_{t + 1}^{\left( k \right)},P_t^c,U_{t + 1}^{\left( k \right)}} \right) \ge \Upsilon \left( {P_{t + 1}^{\left( k \right)},P_{t + 1}^c,U_{t + 1}^{\left( k \right)}} \right).
\end{array}
\end{equation}

This inequation illustrates that the objective function value of problem (4) is monotonically decreasing during the iteration. Thus we complete the proof.

Meanwhile, we can find that problem (4) is lower bounded because it at least satisfies the condition $\Upsilon \left( {P_t^{\left( k \right)},P_t^c,U_t^{\left( k \right)}} \right) \ge 0$, thus \textbf{Theorem 1} guarantees that the proposed method will finally converge to the local optimal solution after a few iterations.

\section{Experiments and analysis}
\subsection{Experimental settings}
\textbf{Dataset}: (1) \textbf{\emph{Handwritten digit}} dataset \cite{cai2013multi}: The used handwritten digit is composed of 2000 samples from 10 digits, \emph{i.e.}, 0-9. Each sample is represented by two views, in which the one is represented by a feature vector with 240 features obtained by the average of pixels in $2 \times 3$ windows, and the other one is represented by the Fourier coefficient vector with 76 features. (2) \textbf{\emph{BUAA-visnir face dataset (BUAA)}} \cite{huang2012buaa}: Following the experimental settings in \cite{zhao2016incomplete}, we evaluate different methods on the first 10 persons with 90 visual images and 90 near infrared images. Each image was pre-resized to a $10 \times 10$ matrix and then transformed into the vector. (3) \textbf{\emph{Cornell}} dataset \cite{blum1998combining,guo2013convex}: This dataset contains 195 webpages collected from the Cornell University. Webpages in the dataset are partitioned into five classes and each webpage is represented by two views, \emph{i.e.}, the content view and citation view. (4) \textbf{\emph{Caltech101}} dataset \cite{fei2007learning}: The original Caltech101 dataset contains 8677 images from 101 objects. In the experiments, a subset named \textbf{Caltech7} \cite{li2015large}, which is composed of 1474 images from 7 classes, is used to compare different methods. The popular two types of features, \emph{i.e.},GIST and LBP, are extracted from each image as the two views. The above used datasets are briefly summarized in Table 1.

\textbf{Evaluation}: Three well-known matrices, \emph{i.e.}, clustering accuracy (ACC), normalized mutual information (NMI), and purity are chosen to evaluate the performance of different methods \cite{cai2013multi}. For the above datasets, we randomly select the percentage of 10, 30, 50, 70, and 90 samples as the paired samples with all views, and treat the remaining samples as incomplete samples, in which half of samples only have one of the views. All methods are repeatedly performed 5 times and their average values (\%) are reported for comparison.

\textbf{Compared methods}: Following the experimental settings in \cite{shao2015multiple,zhao2016incomplete}, we compare the proposed method with the following baselines. (1) \textbf{BSV} (Best Single View): BSV first fills in the missing views with the average of samples in the corresponding view, and then performs \emph{k}-means on each view separately. Finally, the best clustering result of the two views is reported. (2) \textbf{Concat}: It first fills in all missing views with the average of samples of the corresponding view, and then concatenates all views of each sample into one feature vector, followed by performing \emph{k}-means to obtain the clustering reuslt. (3) \textbf{PVC} \cite{zhi2014partial}. PVC uses the non-negative matrix factorization technique to learn a common latent representation for incomplete multi-view clustering. (4) \textbf{IMG} \cite{zhao2016incomplete}: IMG extends the PVC by embedding the adaptively learned Laplacian graph. (5) Double constrained NMF (\textbf{DCNMF}) \cite{qian2016double}: DCNMF is an extension of PVC, which further introduces a Laplacian graph regularizer into PVC. (6) Graph regularized partial multi-view clustering (\textbf{GPMVC}) \cite{rai2016partial}: GPMVC can be viewed as an improved method to DCNMF, which exploits a scale normalization technique in the consensus representation learning term. \emph{The code of the proposed method is available at: http://www.yongxu.org/lunwen.html.}
\begin{table*}[t!]
\caption{Description of the used benchmark datasets.}
\label{tab:1}
\centering
\begin{tabular}{|c|c|c|c|c|c|}
\hline
Database & Class No. & No. of view & No. of samples & Feature No. of \emph{Vi(1)}/\emph{Vi(2)}\\
\hline
\hline
handwritten digit &	10 & 2 	&2000   & 240/76  \\
BUAA              &	10 & 2  & 90    & 100/100 \\
Cornell           & 5  & 2  & 195   & 195/1703 \\
Caltech7          & 7  & 2  & 1474  & 512/928 \\
\hline
\end{tabular}
\end{table*}
\subsection{Experimental results and analyses}
The clustering results of different methods on the above four datasets are enumerated in Table 2-Table 5 and Fig.2. It is obvious to see that the proposed method can significantly improve the ACC, NMI, and purity. In particular, the proposed method achives nearly 8 percent higher than those of the related methods in terms of the ACC on the BUAA dataset. The good performance strongly validates the effectiveness of the proposed method in handling the IMC tasks. Besides, we can obtain the following observations from the experimental results.

(1) Generally, with the ratio of missing views decreases, the clustering performances of all methods improve obviously. This proves that the complementary information of different views is very useful in multi-view learning.

(2) In most cases, BSV and Concat perform much worse than the other methods. This proves that filling in the missing views with the average of samples of the corresponding view is not a useful approach.

(3) DCNMF, GPMVC and the proposed method perform better than PVC in most cases. Compared with PVC, the other two methods and the proposed method all exploit the local geometric structure of each view to guide the representation learning. Thus, the experimental results prove that the local information of each view contain very useful information, which is beneficial to learn a more compact and discriminative representation. Meanwhile, we can find that our method achieves better performance than DCNMF and GPMVC, which further proves the effectiveness of the proposed novel graph regularization term.

\begin{table*}[t!]
\caption{ACCs/NMIs (\%) of different methods on the handwritten digit dataset.}
\label{tab:1}
\centering
\begin{tabular}{|c|c|c|c|c|c|}
\hline
Method/Rate & 0.1 & 0.3 & 0.5 & 0.7 & 0.9 \\
\hline
\hline
BSV	    &43.08/37.04	&50.46/44.48	&57.39/51.50	&64.44/58.61	&69.29/66.26\\
Concat	&46.01/47.71	&57.46/54.43	&66.45/61.12	&78.64/70.30	&86.63/79.34\\
PVC	    &63.81/55.13	&70.90/60.85	&73.44/64.88	&75.20/68.54	&77.82/72.83\\
IMG	    &69.22/58.04	&75.41/62.38	&76.36/64.91	&77.54/68.21	&81.78/73.57\\
DCNMF   &51.21/54.23	&76.63/65.56	&80.61/74.41	&86.16/78.14	&89.16/80.90\\
GPMVC   &65.60/60.99	&74.04/63.99	&76.94/72.23	&79.06/73.68	&81.08/75.24\\
Ours	&\textbf{72.70/66.48}	&\textbf{79.67/71.28}	&\textbf{86.22/77.27}	&\textbf{88.98/80.48}	&\textbf{90.77/83.55}\\
\hline
\end{tabular}
\end{table*}

\begin{table*}[t!]
\caption{ACCs/NMIs (\%) of different methods on the BUAA dataset.}
\label{tab:1}
\centering
\begin{tabular}{|c|c|c|c|c|c|}
\hline
Method/Rate & 0.1 & 0.3 & 0.5 & 0.7 & 0.9 \\
\hline
\hline
BSV	    &48.33/43.10	&56.96/53.03	&64.26/61.78	&70.81/69.91	&80.16/82.56\\
Concat	&45.62/51.22	&46.61/51.95	&47.46/52.43	&52.34/56.51	&57.58/62.66\\
PVC	    &57.41/61.35	&66.46/67.07	&70.01/71.97	&75.92/78.70	&80.73/84.22\\
IMG	    &53.95/54.72	&67.39/67.53	&76.14/76.74	&79.36/82.83	&80.78/85.90\\
DCNMF   &58.36/61.78	&67.58/68.75	&72.15/72.05	&76.58/79.66	&82.42/86.42\\
GPMVC   &58.98/62.12	&68.75/70.25	&74.28/74.33	&78.28/81.63	&84.24/86.78\\
Ours	&\textbf{63.82/64.64}	&\textbf{76.72/76.04}	&\textbf{82.76/81.35}	&\textbf{86.20/85.77}	&\textbf{92.62/91.20}\\
\hline
\end{tabular}
\end{table*}
\begin{table*}[t!]
\caption{ACCs/NMIs (\%) of different methods on the cornell dataset.}
\label{tab:1}
\centering
\begin{tabular}{|c|c|c|c|c|c|}
\hline
Method/Rate & 0.1 & 0.3 & 0.5 & 0.7 & 0.9 \\
\hline
\hline
BSV	    &42.41/8.66	    &43.93/8.19  	&44.84/8.89 	&46.32/12.69	&47.66/19.34\\
Concat	&38.80/8.07	    &38.06/7.56	    &36.96/8.30	    &36.79/10.21	&38.48/13.47\\
PVC	    &42.56/15.76	&42.56/16.00	&43.79/18.21	&42.56/19.76	&43.03/21.03\\
IMG	    &45.13/12.56	&45.79/16.62	&47.08/19.24	&45.51/20.89	&44.76/22.98\\
DCNMF   &39.94/13.59	&43.29/17.72	&43.18/19.17	&45.74/21.69	&45.52/23.98\\
GPMVC   &40.39/13.90	&43.86/16.07	&46.53/18.99	&44.56/15.03	&44.35/17.07\\
Ours	&\textbf{46.99/17.23}	&\textbf{47.40/18.36}	&\textbf{49.03/21.01}	&\textbf{48.78/22.11}	&\textbf{49.20/25.02}\\
\hline
\end{tabular}
\end{table*}
\begin{table*}[t!]
\caption{ACCs/NMIs (\%) of different methods on the Caltech7 dataset.}
\label{tab:1}
\centering
\begin{tabular}{|c|c|c|c|c|c|}
\hline
Method/Rate & 0.1 & 0.3 & 0.5 & 0.7 & 0.9 \\
\hline
\hline
BSV	    &42.66/29.04	&40.97/32.11	&39.83/35.13	&42.92/38.83	&46.99/44.16\\
Concat	&36.83/33.82	&31.74/34.44	&36.36/34.56	&43.38/38.15	&47.08/45.44\\
PVC	    &43.46/38.99	&43.96/40.26	&44.46/40.17	&44.76/41.60	&44.34/41.94\\
IMG	    &42.05/32.38	&42.36/33.29	&42.23/35.05	&41.17/35.96	&43.23/37.64\\
DCNMF   &40.63/33.86	&44.53/38.19	&45.62/41.40	&48.50/41.24	&50.74/44.04\\
GPMVC   &45.57/40.05	&47.19/40.96	&46.99/41.83	&46.99/42.61	&49.10/46.02\\
Ours	&\textbf{50.88/41.01}	&\textbf{51.15/42.74}	&\textbf{51.40/45.69}	&\textbf{51.79/46.78}	&\textbf{51.88/48.28}\\
\hline
\end{tabular}
\end{table*}
\begin{figure}[t!]
\centering
\subfloat[Handwritten digit]{\includegraphics[width=2in,height=1.2in]{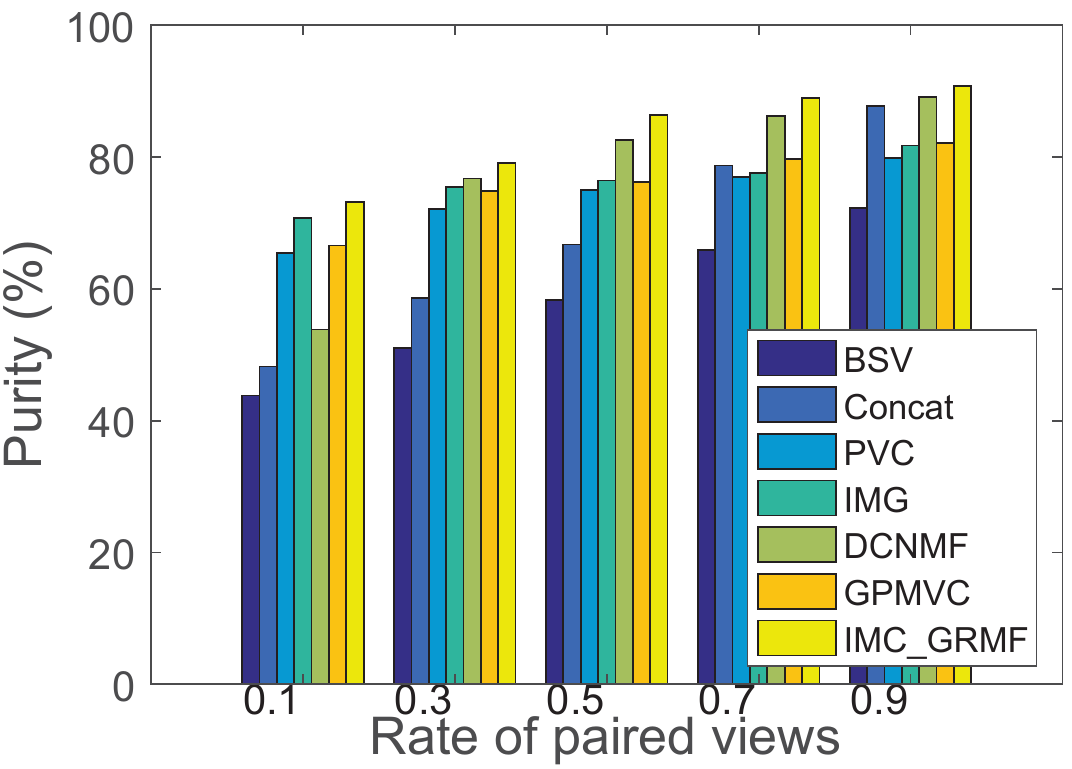}
\label{fig_first_case}}
\hfil
\subfloat[BUAA]{\includegraphics[width=2in,height=1.2in]{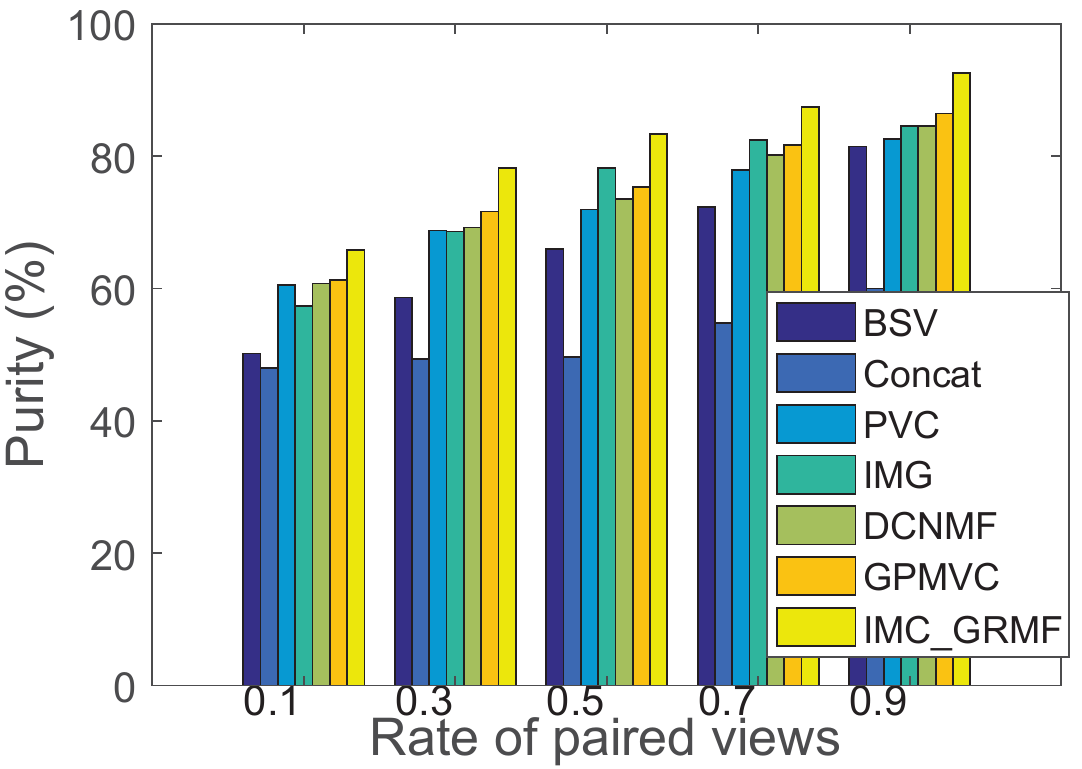}
\label{fig_second_case}}
\hfil
\subfloat[Cornell]{\includegraphics[width=2in,height=1.2in]{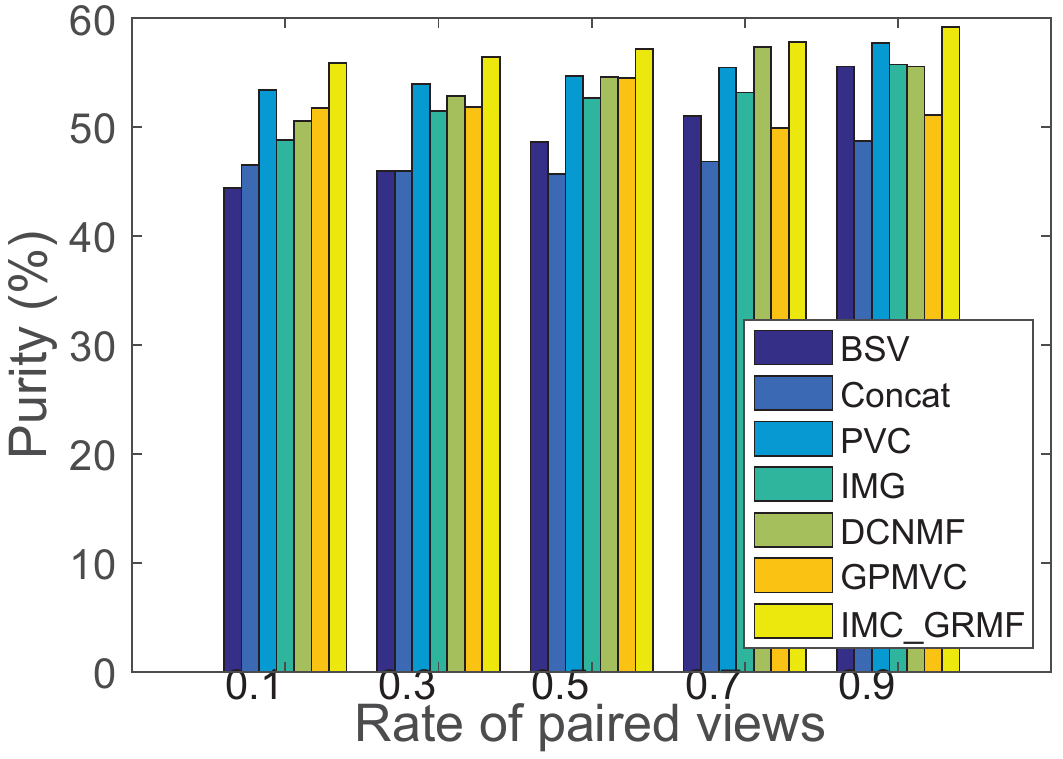}
\label{fig_second_case}}
\hfil
\subfloat[Caltech7]{\includegraphics[width=2in,height=1.2in]{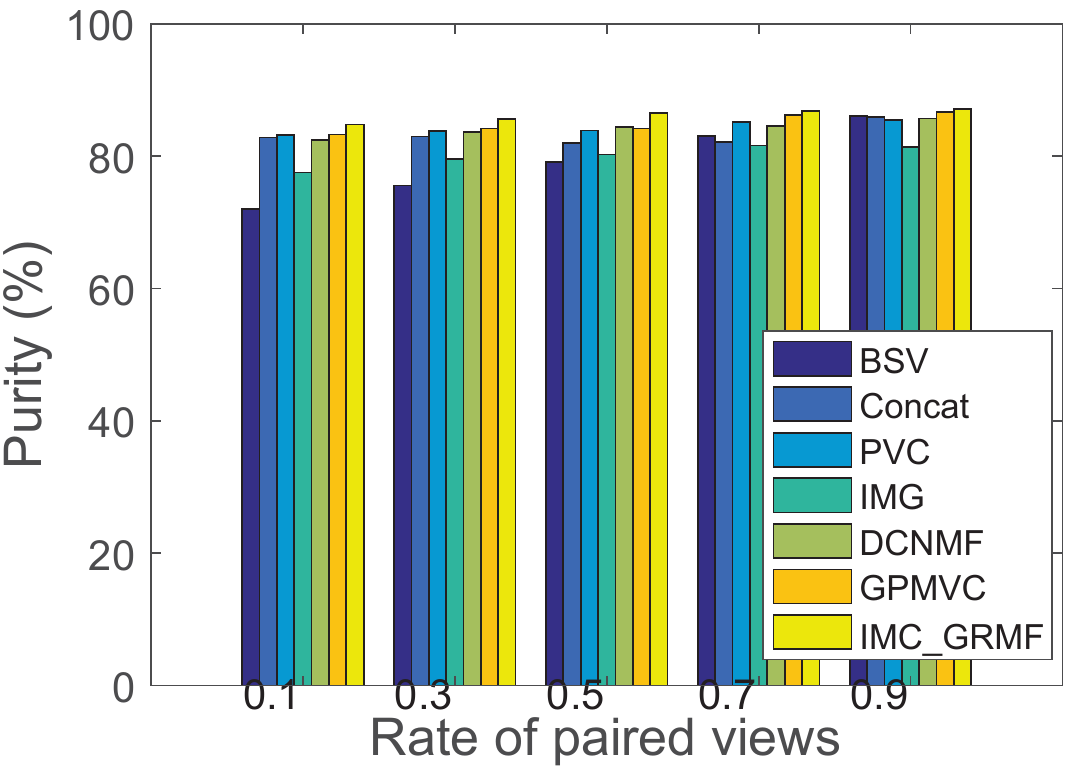}
\label{fig_second_case}}
\hfil
\caption{Purity (\%) of different methods on the above four datasets.}
\label{fig:3}
\end{figure}
\subsection{Parameter analysis}
Fig.3 shows the ACC versus the parameters ${\lambda _1}$ and ${\lambda _2}$ on the handwritten digit and BUAA datasets with 70\% paired samples. It is obvious that the ACC of the proposed method is relatively stable in some local areas, which indicates that the proposed method is insensitive to the selection of parameters to some extent. Moreover, we can find that when the two parameters are selected with proper values from the candidate range of ($[ {{{10}^0},{{10}^2}}]$, $[ {{{10}^{{\rm{ - 5}}}},{{10}^{{\rm{ - 1}}}}}]$), the proposed method can achieve the satisfactory performance. This indicates that a relative larger value of parameter $\lambda_1$ encourages a better performance. In our work, we use the grid searching approach to find the optimal combinations of the two parameters from the two dimensional grid formed by ($[ {{{10}^0},{{10}^2}}]$, $[ {{{10}^{{\rm{ - 5}}}},{{10}^{{\rm{ - 1}}}}}]$) \cite{zhang2017elastic}.

Fig.4 plots the relationships of ACC and the number of nearest neighbors of the proposed method on the handwritten digit and BUAA datasets. From the figures, we have the following conclusions: (1) The clustering performance is insensitive to the selection of nearest neighbor number to some extent when the nearest neighbor number is located in the proper range, such as $\left[ {8,18} \right]$ for the handwritten digit dataset and $\left[ {2,6} \right]$ for the BUAA dataset. (2) Generally, the number of nearest neighbors should better be less than the number of sample of each class. For example, from Fig.3(b), we can find that when the number of nearest neighbors is larger than the number of sample per class, \emph{i.e.}, $N > 10$, the ACC decreases dramatically. However, in the real world applications, it is impossible to obtain the real number of sample per class. In this work, we use the following criterion to select the number of the nearest neighbors. Suppose we try to partition the available multi-view data with $n$ samples into $c$ groups, $m = {n \mathord{\left/
 {\vphantom {n c}} \right.\kern-\nulldelimiterspace} c}$. If $m \gg 10$, then we empirically select 10 as the number of nearest neighbors, otherwise we select $min(m - 4,2)$ as the nearest neighbor number.
\begin{figure}[t!]
\centering
\subfloat[Handwritten digit]{\includegraphics[width=2in,height=1.2in]{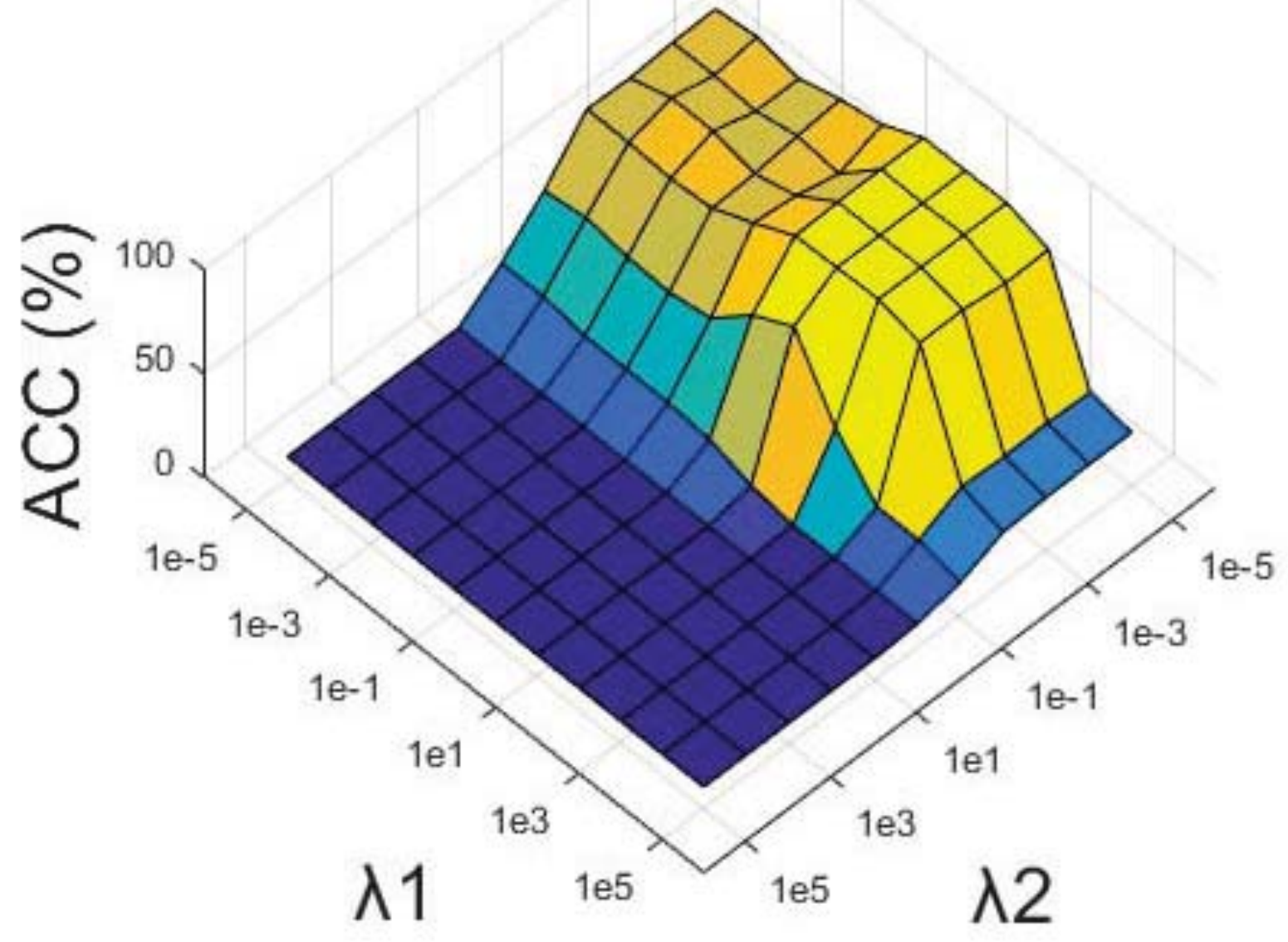}
\label{fig_first_case}}
\hfil
\subfloat[BUAA]{\includegraphics[width=2in,height=1.2in]{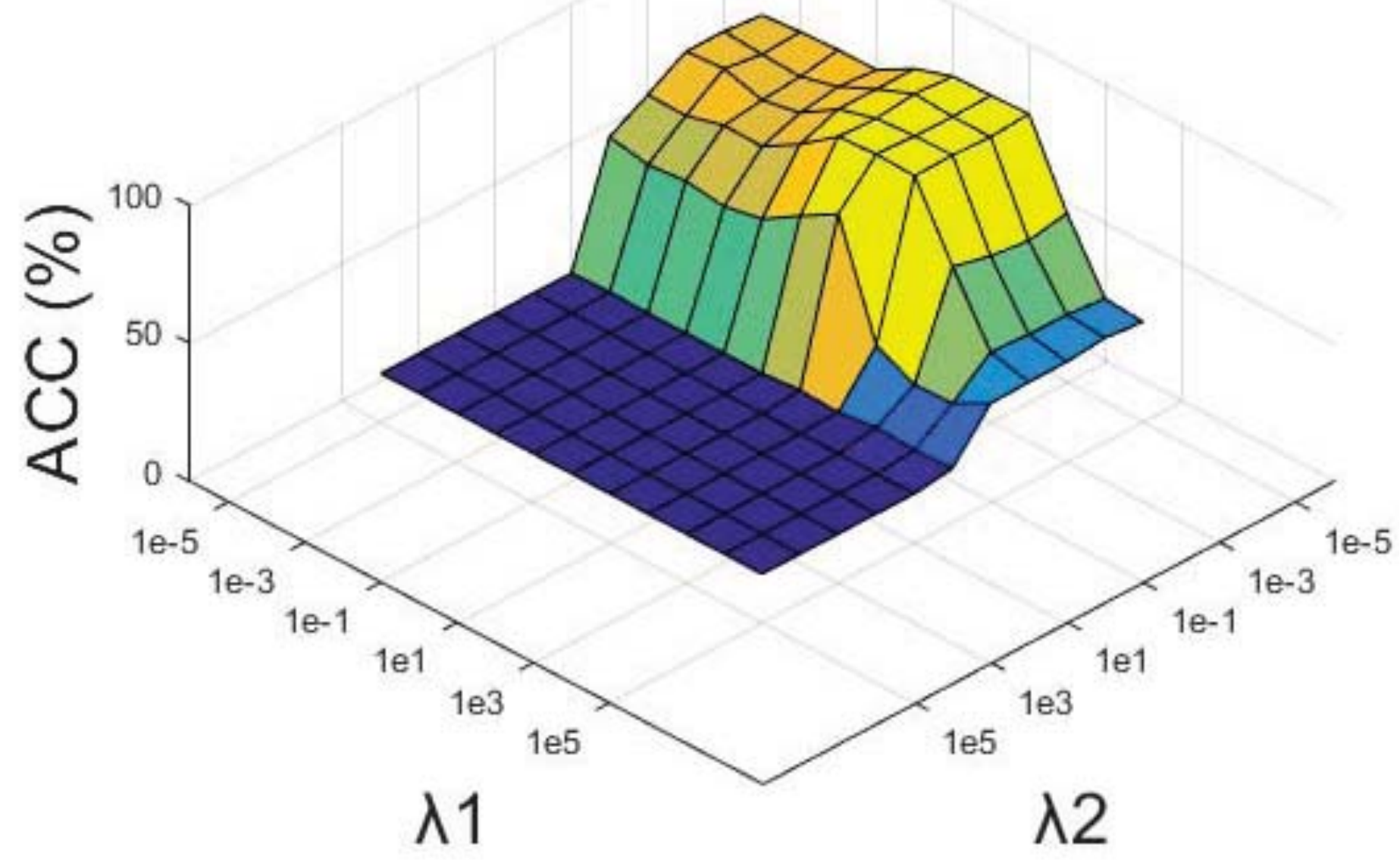}
\label{fig_second_case}}
\caption{ACC (\%) versus parameters ${\lambda _{\rm{1}}}$ and ${\lambda _{\rm{2}}}$ of the proposed method on (a) handwritten digit and (b) BUAA datasets with 70\% paired samples.}
\label{fig:3}
\end{figure}
\begin{figure}[t!]
\centering
\subfloat[Handwritten digit]{\includegraphics[width=2in,height=1.2in]{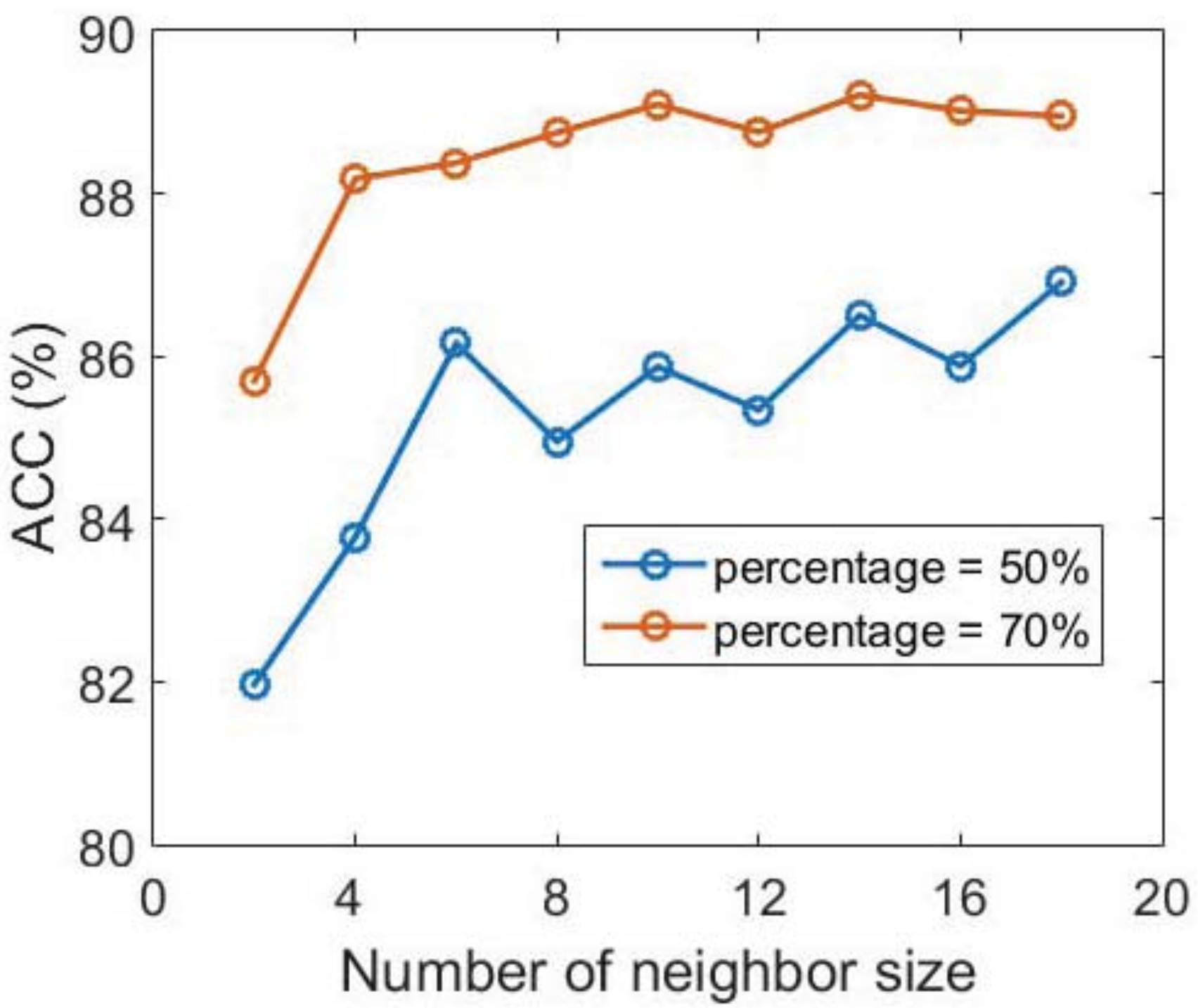}
\label{fig_first_case}}
\hfil
\subfloat[BUAA]{\includegraphics[width=2in,height=1.2in]{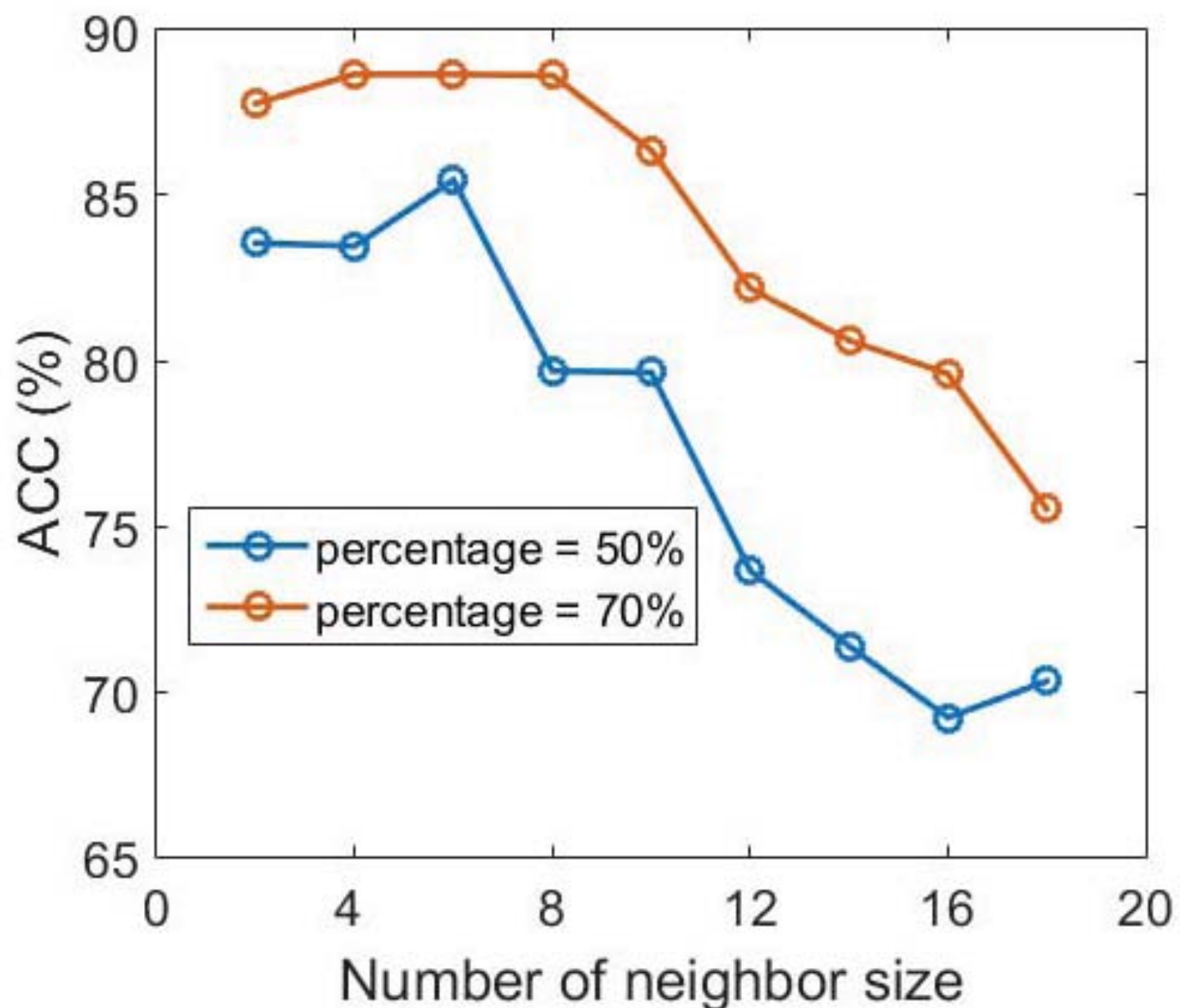}
\label{fig_second_case}}
\caption{ACC (\%) versus the number of nearest neighbors of our method on (a) handwritten digit and (b) BUAA datasets with 50\% and 70\% paired samples.}
\label{fig:3}
\end{figure}
\subsection{Experimental convergence study}
Fig.5 shows the objective function value and ACC at each iteration step on the handwritten digit and BUAA datasets with 70\% paired samples. From the figures, it is obvious to see that the objective function value decreases dramatically in the first few iteration steps (within 20 iterations). The experimental results plotted in the two figures prove the good convergence property of our method.
\begin{figure}[t!]
\centering
\subfloat[Handwritten digit]{\includegraphics[width=2in,height=1in]{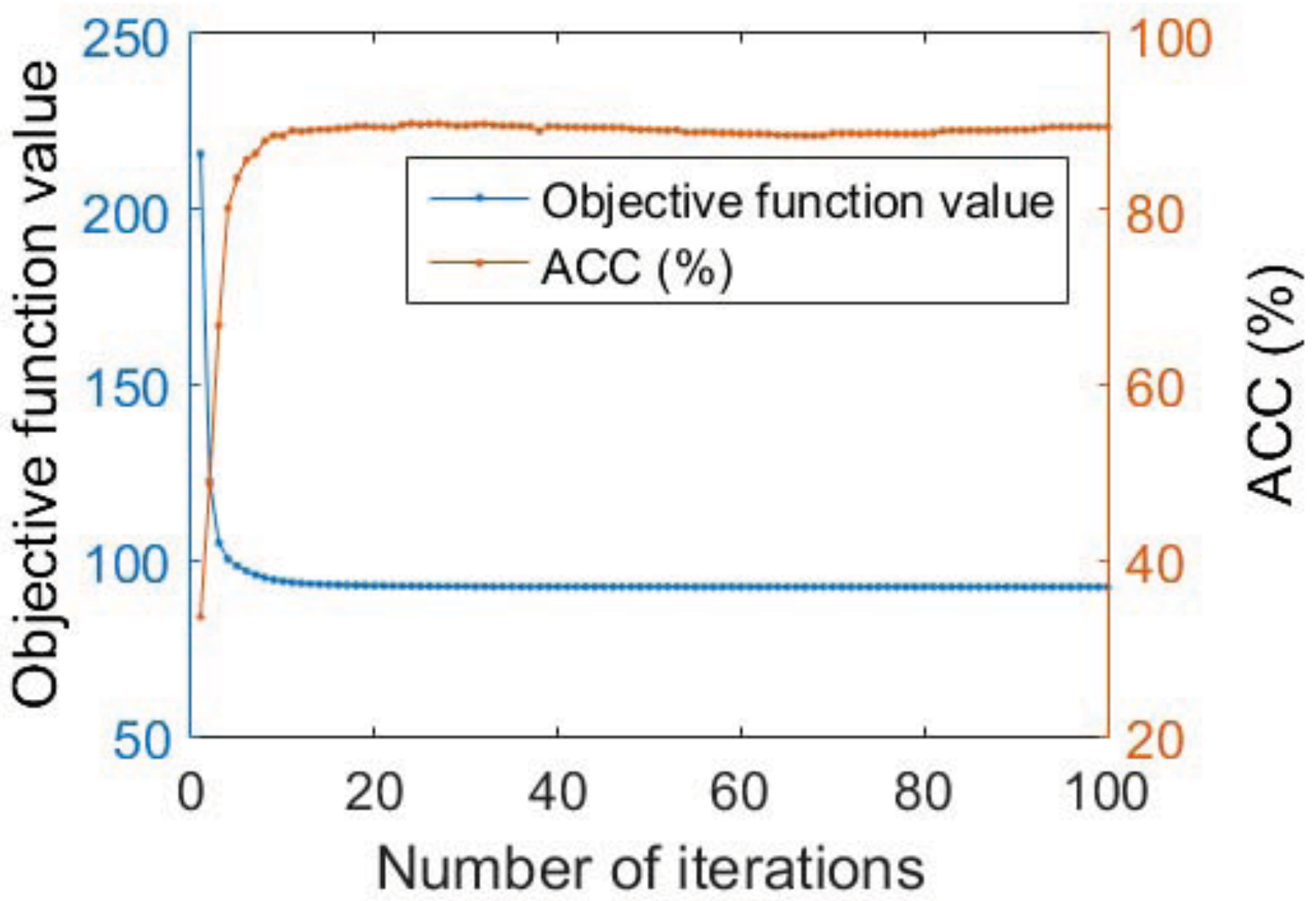}
\label{fig_first_case}}
\hfil
\subfloat[BUAA]{\includegraphics[width=2in,height=1in]{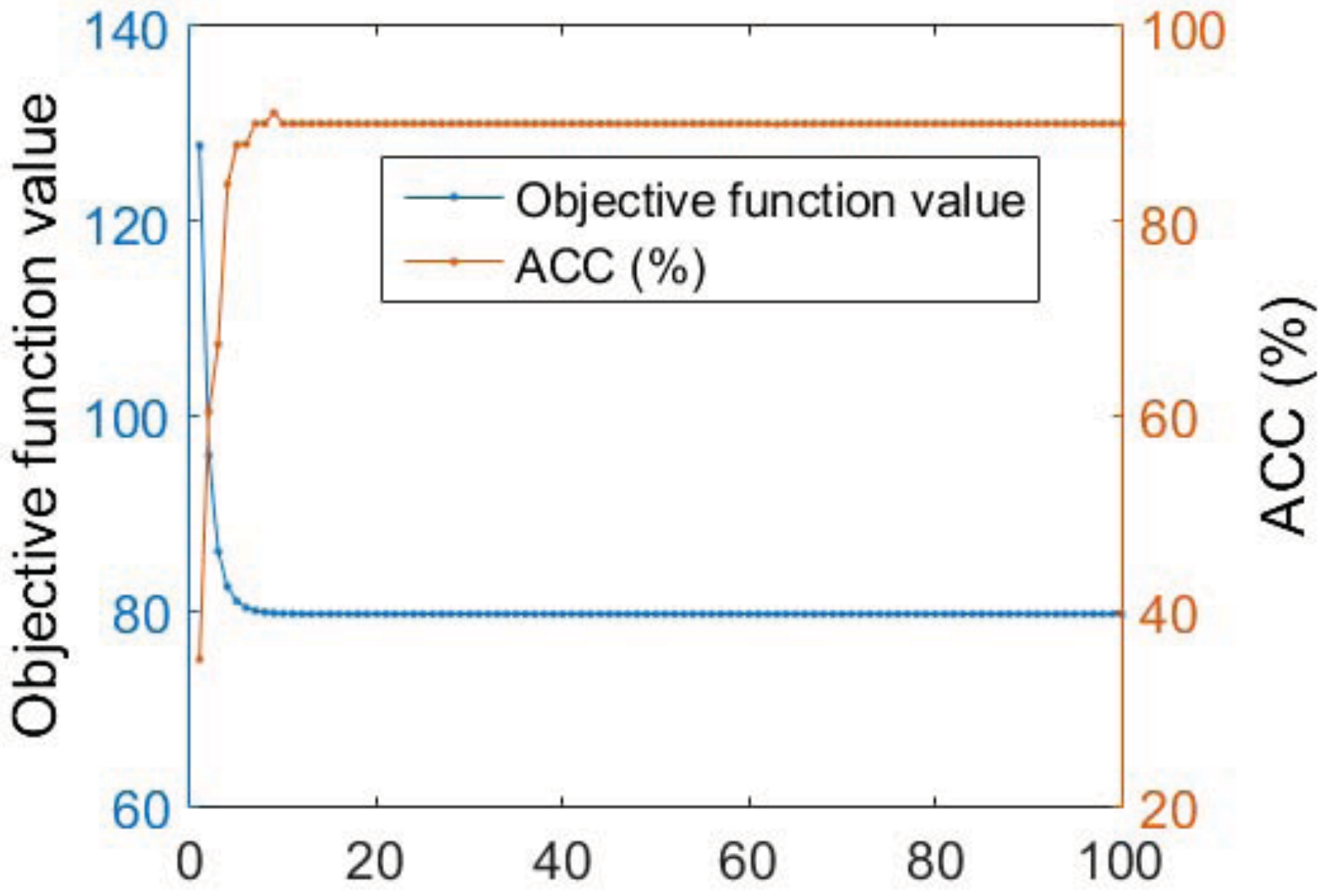}
\label{fig_second_case}}
\caption{The objective function value and ACC (\%) versus the iteration step of the proposed method on (a) handwritten digit and (b) BUAA datasets with 70\% paired samples.}
\label{fig:3}
\end{figure}
\section{Conclusions}
In this paper, we propose a novel framework for multi-view learning, which not only can handle the incomplete and complete multi-view clustering, but also is able to deal with the out-of-sample. Moreover, the proposed method has the potential to complete the missing views for any sample. Besides, we provide a novel approach to exploit the local information of data without introducing any extra regularization term and penalty parameter, which does not increase the complexity and computational burden. Extensive experimental results prove the effectiveness of the proposed method.
\section{Acknowledgments}
This work is supported in part by Economic, Trade and information Commission of Shenzhen Municipality (Grant no. 20170504160426188).

\bibliographystyle{splncs04}
\bibliography{egbib}
\end{document}